\documentclass[sigconf]{acmart}
\usepackage{balance}
\usepackage{xspace}
\makeatletter
\DeclareRobustCommand\onedot{\futurelet\@let@token\@onedot}
\def\@onedot{\ifx\@let@token.\else.\null\fi\xspace}
 
\def\ie{{i.e}\onedot}

\makeatother

\renewcommand\footnotetextcopyrightpermission[1]{} 
\AtBeginDocument{%
  \providecommand\BibTeX{{%
    \normalfont B\kern-0.5em{\scshape i\kern-0.25em b}\kern-0.8em\TeX}}}

\setcopyright{acmcopyright}
\copyrightyear{2021}
\acmYear{2021}
\acmDOI{10.1145/1122445.1122456}

\begin{document}

\title{SOGAN: 3D-Aware Shadow and Occlusion Robust GAN for Makeup Transfer}
\fancyhead{}


\author{Yueming Lyu$^{1,2}$, Jing Dong$^{1*}$, Bo Peng$^{1,3}$, Wei Wang$^{1}$, Tieniu Tan$^{1}$  \\
$^{1}$ CRIPAC, NLPR \\
Institute of Automation, Chinese Academy of Sciences (CASIA) \\
$^{2}$ School of Artificial Intelligence, University of Chinese Academy of Sciences (UCAS)\\
$^{3}$ State Key Laboratory of Information Security, IIE, CAS \\
{\tt\small yueming.lv@cripac.ia.ac.cn, \{jdong, bo.peng, wwang, tnt\}@nlpr.ia.ac.cn}}





\renewcommand{\shortauthors}{Lyu et al.}

\begin{abstract} 
	\vspace{1ex}
  In recent years, virtual makeup applications have become more and more popular.
  However, it is still challenging to propose a robust makeup transfer method in the real-world environment.  
  Current makeup transfer methods mostly work well on good-conditioned clean makeup images, but transferring makeup that exhibits shadow and occlusion is not satisfying. 
  To alleviate it, we propose a novel makeup transfer method, called 3D-Aware Shadow and Occlusion Robust GAN (SOGAN).
  Given the source and the reference faces, we first fit a 3D face model and then disentangle the faces into shape and texture. 
  In the texture branch, we map the texture to the UV space and design a UV texture generator to transfer the makeup. 
  Since human faces are symmetrical in the UV space, we can conveniently remove the undesired shadow and occlusion from the reference image by carefully designing a Flip Attention Module (FAM). 
  After obtaining cleaner makeup features from the reference image, a Makeup Transfer Module (MTM) is introduced to perform accurate makeup transfer. 
  The qualitative and quantitative experiments demonstrate that our SOGAN not only achieves superior results in shadow and occlusion situations but also performs well in large pose and expression variations.
\end{abstract}

\maketitle

\begin{figure}[t]
	\begin{center}
		\includegraphics[width=1.0\linewidth]{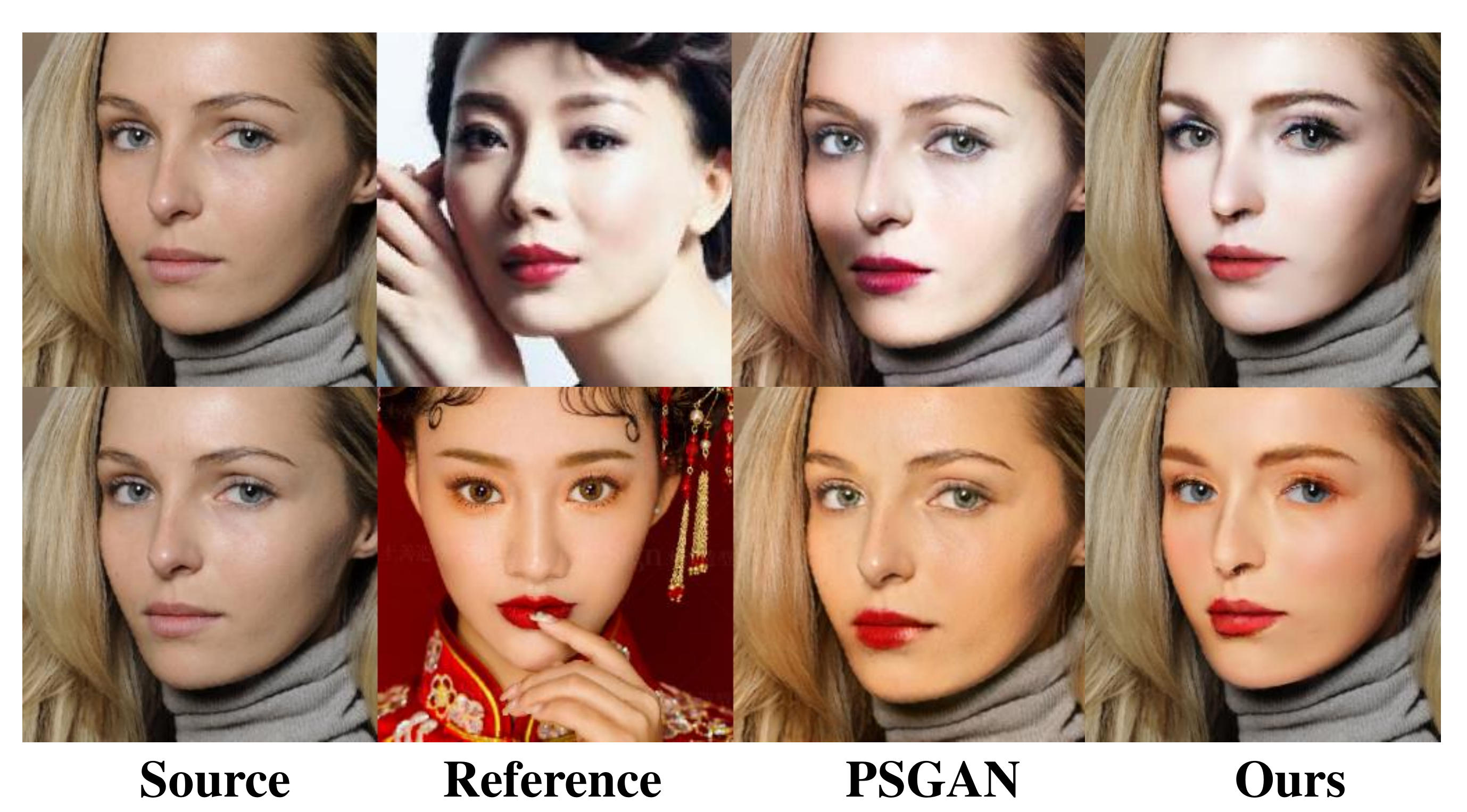}
	\end{center}
	\caption{Comparisons with the current state-of-the-art method in shadow and occlusion situations. From the left to the right, we show the input source images, the input reference images, the results of PSGAN \cite{jiang2020psgan}, and the results of our method.}
	\label{fig:the first}
\end{figure}
\section{Introduction}

Facial makeup transfer aims at transferring the makeup style from a reference makeup face to a non-makeup face without changing the identity and the background. 
Recently, there is strong demand for portrait beautification and more and more virtual makeup applications appear in the public. 
Most current methods transfer the makeup based on the architecture of CycleGAN \cite{zhu2017unpaired} in the 2D image space.
For example, PairedCycleGAN \cite{chang2018pairedcyclegan}trains three generators and discriminators separately for each facial component. 
BeautyGAN \cite{li2018beautygan} introduces both global domain-level loss and local instance-level loss in a dual input/output generator. 
In particular, PSGAN \cite{jiang2020psgan} achieves state-of-the-art results even under large pose and expression variations.
\vspace{1ex}

However, current 2D-based methods have some limitations. 
They mostly work well on completely clean makeup reference images.
As shown in the first row of Figure~\ref{fig:the first}, 
when there is shadow contaminating the makeup area, even the state-of-the-art method, PSGAN \cite{jiang2020psgan}, mistakenly transfers the shadow and leads to ghosting artifacts in the output image. 
In addition, as shown in the second row of Figure~\ref{fig:the first}, when there is occlusion in the makeup area of the reference image, such as hands covering the lips, current method fails to repair the occluded makeup regions. 
Consequently, it produces a result with missing makeup details in the corresponding position of the output image. 
The poor makeup transfer caused by these contaminated regions is hard to handle in the 2D space since shadow and occlusion are natural effects from 3D spatial relationships between facial components and objects.
However, a practically applicable makeup transfer method should be shadow and occlusion robust, which is able to generate high-quality results even if reference images show contaminated makeup styles.

\begin{figure}[t]
	\begin{center}
		\includegraphics[width=1.0\linewidth]{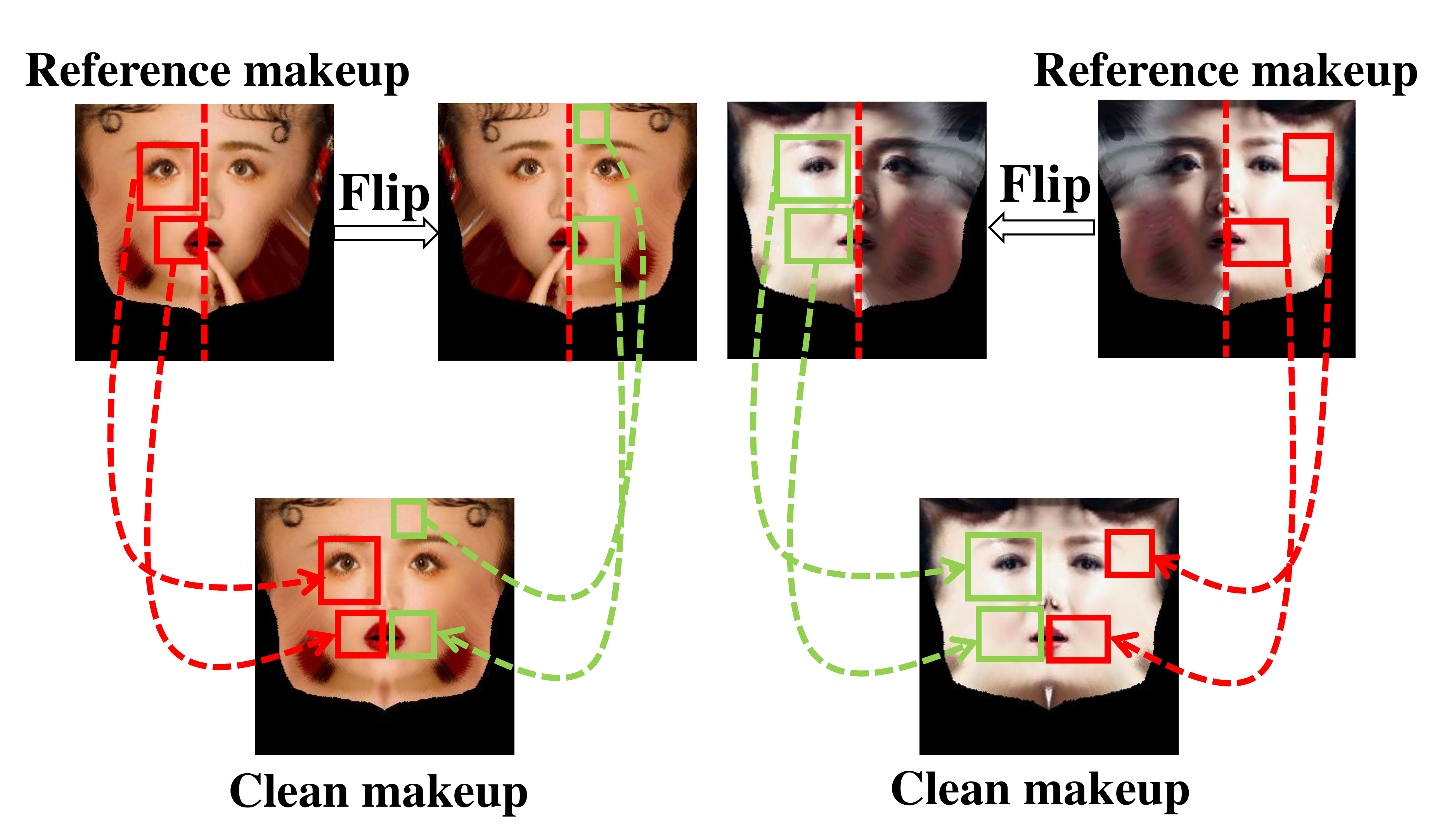}
	\end{center}
	\caption{An illustration of our basic idea on tackling the shadow and occlusion problem.
	In the UV space, our method repairs the contaminated regions of the reference makeup by adaptively extracting clean makeup information from both original and flipped makeup textures (indicated by red and green).  
	Note that our method learns cleaner makeup representation in the feature space in our actual model.}
	\label{fig:uv_texture}
\end{figure}

\vspace{1ex}

To address the above issues, this paper proposes a novel 3D-Aware Shadow and Occlusion Robust GAN (SOGAN) for makeup transfer.
We first obtain the shape and texture of input images by fitting a 3D face model and then map the texture to the UV space. 
Inspired by \cite{Wu_2020_CVPR} that exploits the underlying object symmetry to reconstruct 3D objects, we leverage the bilateral symmetry of faces in the UV space to adaptively repair the contaminated facial makeup (as shown in Figure~\ref{fig:uv_texture}).
Specifically, we design a UV texture generator to transfer the makeup in the texture space while keeping the shape unchanged.
With the source texture and the reference makeup texture as inputs, the UV texture generator can produce a new texture that corresponds to the desired makeup style. 
It consists of two modules: Flip Attention Module (FAM) and Makeup Transfer Module (MTM).
The FAM is proposed to adaptively grasp the clean regions of the reference image and repair the shadowed and occluded regions by applying flipping operation and attention fusion.  
After obtaining cleaner makeup representation, MTM is introduced to transfer makeup accurately. 
Different from previous methods that transfer makeup by building complex pixel-wise correspondence between 2D faces, 
MTM directly proposes an attention map from the source to the reference and obtains the makeup details of the corresponding locations.
Since each position in the UV space has the same semantic meaning, it is easier to transfer the most related regions between faces compared to 2D-based methods. 
In addition, due to the pose and expression variations are explicitly normalized in the UV space,
our method naturally achieves impressive transfer results under large pose and expression.
Finally, we take advantage of the differentiable rendering method to propagate the gradients back to the whole network from the rendering output. 
In this way, our method can be trained in an end-to-end manner. 
\vspace{1ex}

We summarize our contributions as follows:
\begin{itemize}
	\item We propose a 3D-aware makeup transfer method to disentangle the shape and texture of input images and only transfer the makeup in the UV texture space. 
	\item We utilize the symmetry of faces in the UV space and propose a Flip Attention Module (FAM) and a Makeup Transfer Module (MTM) to alleviate the influence of shadow and occlusion. 
	\item The qualitative and quantitative experiments demonstrate that our SOGAN not only achieves state-of-the-art results in shadow and occlusion situations but also performs well in large pose and expression variations.
\end{itemize}

\begin{figure*}
	\begin{center}
		\includegraphics[height=0.6\textheight]{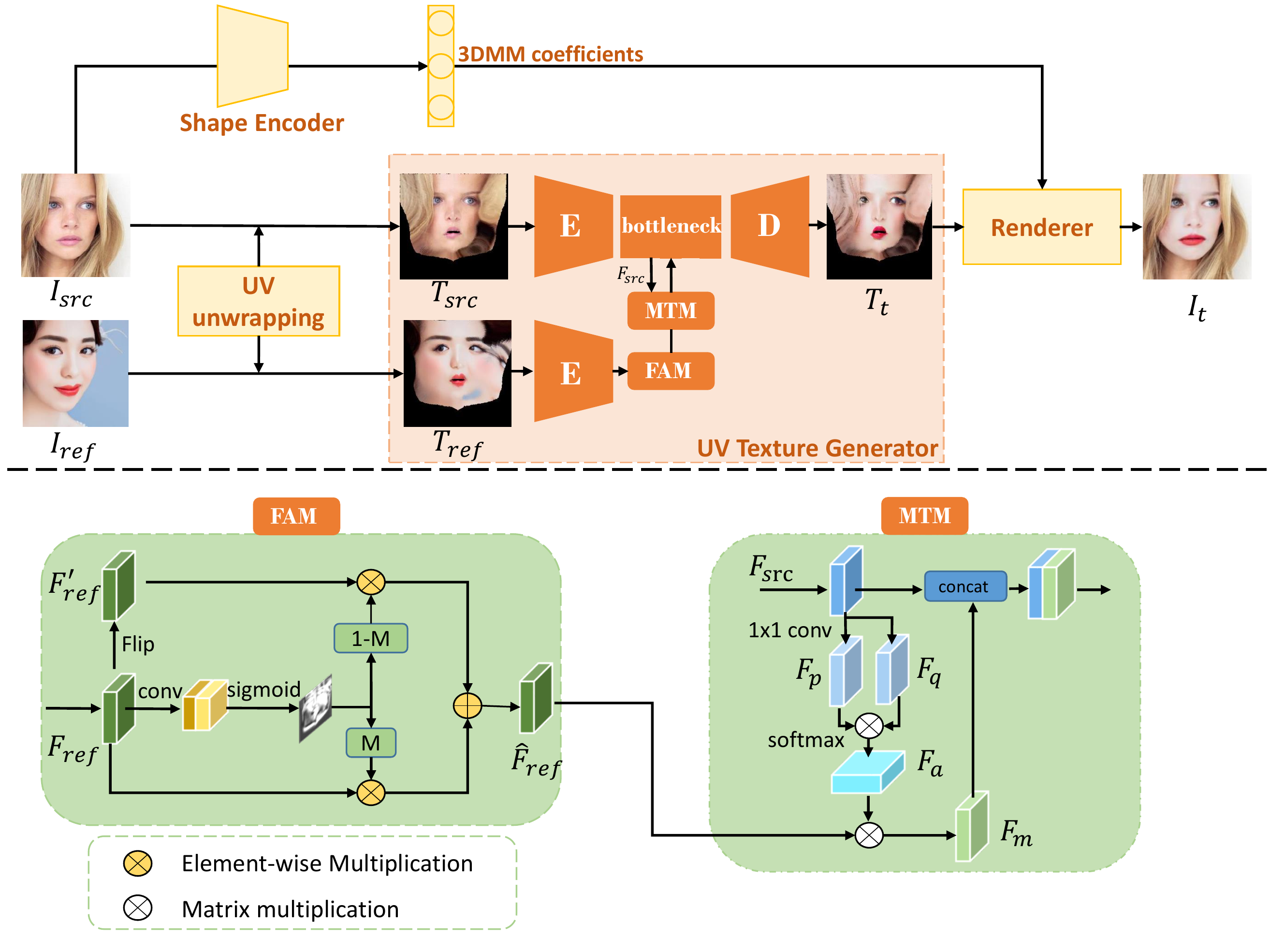}
	\end{center}
	\vspace{-1ex}
	\caption{An overview of the architecture of our method. The top figure is the overall pipeline of our SOGAN. The bottom-left figure shows details of the proposed FAM module, which is used to repair the shadowed and occluded regions by applying flipping operation and attention fusion. The bottom-right figure shows details of the proposed MTM module, which is responsible for accurate makeup transfer.}
	\label{fig:pipeline}
\end{figure*}

\section{related work}
\subsection{Makeup Transfer}
Makeup transfer aims at transferring the makeup from a reference face image to the source non-makeup one while preserving the source identity.
As a fundamental work in image-to-image translation, CycleGAN \cite{zhu2017unpaired} can be directly exploited to makeup transfer by introducing two coupled generators and a cycle consistency loss. 
However, it can only transfer a general makeup style, rather than specific makeup details.
Based on CycleGAN\cite{zhu2017unpaired}, PairedCycleGAN \cite{chang2018pairedcyclegan} designs three local generators and discriminators for eyes, lips, and skin regions and then uses Poisson Blending to blend the three regions into the source image.
Similar to PairedCycleGAN\cite{chang2018pairedcyclegan}, LADN \cite{gu2019ladn} designs multiple local discriminators on overlapping image regions to better transfer local details.
BeautyGAN\cite{li2018beautygan} proposes a dual input/output generator which performs makeup transfer and makeup removal in a unified network.
Furthermore, they introduce a makeup loss by calculating pixel-level histogram differences between the reference face image and the output one.
BeautyGlow \cite{chen2019beautyglow}, based on the Glow architecture, decomposes the latent vectors into makeup and non-makeup latent vectors.
Recently, PSGAN \cite{jiang2020psgan} proposes an attentive makeup morphing module to better transfer makeup under large pose and expression situations. 
This is one step forward from the restricted frontal faces to the faces in the wild.
In order to further push makeup transfer towards real-world usage, we try to solve the problem of undesired shadow and occlusion in the reference images. 
To achieve this goal, we propose a novel 3D-aware method to repair shadowed and occluded regions in the reference makeup images.
\subsection{3D Face Reconstruction}
3D face reconstruction aims to recover the 3D shape and texture of human faces from 2D images. 
The widely-used method is 3D Morphable Models (3DMM) \cite{blanz1999morphable,paysan20093d,blanz2003reanimating,blanz2003face}. 
It is based on the combination of parametric descriptions of 3D face geometry and texture with PCAs, which are built from a collection of real scans. 
With the development of deep learning, some methods use CNN to estimate the 3D Morphable Model parameters \cite{zhu2016face,guo2020towards}.
Using 3DMM parameters, we can obtain a series of 3D representations, such as 3D face shape, texture, and UV texture map \cite{feng2018joint,guo2020beyond}. 
In addition, neural renderer \cite{kim2018deep,thies2019deferred,gecer2018semi} can be used to fit 3D face models to 2D images.
Recently, \citet{Wu_2020_CVPR} exploits the underlying object symmetry (e.g.\ faces) to deform 3D objects in an unsupervised manner.
In this paper, we first exploit 3DMM to obtain the shape and texture representation and then exploit symmetry to perform makeup transfer on the UV texture map, inspired by \cite{Wu_2020_CVPR}. 
Finally, we render it to the 2D space to get the makeup transfer image. 
\subsection{3D-Aware Face Synthesis}
Face synthesis\cite{li2020dual,deng2020disentangled,fu2021dvg,fu2019dual,li2019global,li2020hierarchical} has achieved rapid progress with the help of deep learning.
In recent years, many methods are proposed to apply 3D priors into GAN schemes for face synthesis \cite{gecer2018semi,deng2020disentangled,tewari2020stylerig,yin2017towards,geng20193d,nagano2018pagan,thies2016face2face,xu2020deep,kim2018deep,hu2020face,ren2019face}.
Most of them leverage 3DMMs.
For face manipulation,  
\citet{gecer2018semi} translates rendered 3D faces and real faces in a cycle fashion to produce new identities with wide ranges of poses, expressions, and illuminations. 
\citet{deng2020disentangled} embeds 3D priors in an imitative-contrastive learning scheme and achieve satisfactory disentanglement of attributes.
\citet{tewari2020stylerig} provides a face rig-like control over a pre-trained StyleGAN via a 3DMM in a self-supervised manner.
For frontal face synthesis, \citet{yin2017towards} estimates the 3DMM coefficients of input non-frontal faces as low-frequency features to synthesize frontal faces. 
For facial expression editing, \citet{geng20193d} leverages a 3D face model on the input image to disentangle texture and shape spaces, and then apply transformations in these two spaces.
\citet{nagano2018pagan} allows fine-grained control over facial expression by using a combination of rendered expressions, normal maps, and eye-gaze indicators.
For face swapping or reenactment, based on underlying 3D face representations, many methods provide control over facial appearances.
\citet{thies2016face2face} fits a 3D face model to both faces and then transfers the expression components of one face onto the other. 
\citet{xu2020deep} proposes a two-step geometry learning scheme and provides high-fidelity 3D head geometry and head pose manipulation results.
\citet{kim2018deep} proposes a space-time architecture to transfer the full head from a source face to a portrait video of a target face. 
For face deblurring and super-resolution, \citet{hu2020face} incorporates 3D priors into their network and obtain impressive results.
\citet{ren2019face} leverages both image intensity and high-level identity information derived from the reconstructed 3D faces to deblur the input face video. 
Similar to these methods, we also employ 3DMM into our makeup transfer pipeline.
\section{Method}
Our pipeline is shown in Figure~\ref{fig:pipeline}.
Our goal is to transfer the makeup style from the reference image $I_{ref}$ to the source image $I_{src}$, while preserving the identity and background of the source image $I_{src}$. 
We first fit the 3D face shape and camera projection matrix from both images, with which we extract UV textures of the source and the reference (subsection~\ref{subsection:3d}). 
Then we design a UV texture generator to generate the transferred texture containing the makeup details of the reference image $I_{ref}$ (subsection~\ref{subsection:texG}).
The transferred texture with reference makeup style can be formulated as $T_{t}=G(T_{src},T_{ref})$, where $T_{src}$ is the UV texture of $I_{src}$, $T_{ref}$ is the UV texture of $I_{ref}$. 
After obtaining the transferred texture $T_{t}$, we fed it with the source shape to render a transferred image $I_{t}$.
It is expressed as $I_{t}=R(S_{src}, G(T_{src},T_{ref}))$, where R is a differentiable renderer.
Within the rendering loop, we introduce four types of loss functions, \ie the makeup loss, the perceptual loss, the cycle consistency loss, and the adversarial loss, to guide the process of makeup transfer (subsection~\ref{subsection:loss}).  


\subsection{3D Face Fitting}
\label{subsection:3d}
\citet{blanz2003face} propose the 3D Morphable Model (3DMM) to describe 3D face with PCAs.
With the development of deep learning, some methods use CNN to estimate the 3DMM coefficients \cite{zhu2016face,guo2020towards}. 
Here, we apply 3DDFA \cite{zhu2016face} to estimate the 3DMM coefficients.
We can construct a 3D face shape $\mathbf{S}$ and texture $\mathbf{T}$ for the input faces as:

\begin{equation}
  \begin{aligned}
  \mathbf{S} &=\overline{\mathbf{S}}+\mathbf{A}_{id} \alpha_{id}+\mathbf{A}_{\exp} \alpha_{\exp } \text{,}\\
  \mathbf{T} &=\overline{\mathbf{T}}+\mathbf{A}_{tex} \alpha_{tex} \text{,}
  \end{aligned} 
\end{equation}

where $\overline{\mathbf{S}}$ is the mean face shape, $\mathbf{A}_{id}$ and $\mathbf{A}_{\exp}$ are the PCA bases of identity and expression, respectively. 
The 3D shape coordinates $\mathbf{S}$ are computed as the linear combination of $\overline{\mathbf{S}}$, $\mathbf{A}_{id}$, and $\mathbf{A}_{\exp}$.
While $\overline{\mathbf{T}}$ is the mean face texture, $\mathbf{A}_{tex}$ is the PCA bases of texture. 
The texture $\mathbf{T}$ can be obtained by the linear combination of $\overline{\mathbf{T}}$ and $\mathbf{A}_{tex}$. \\

In particular, the texture $\mathbf{T}$ is parameterized as a vector, which contains the texture intensity value of each vertex in the 3D face mesh. 
In order to obtain the UV texture representation, we project each 3D vertex onto the UV space using cylindrical unwrap. 
In this way, the UV texture can be extracted with the UV coordinates, camera projection, and fitted 3D shape. 
\subsection{UV Texture Generator}
\label{subsection:texG}
We model the makeup transfer task as a problem that keeps the shape of the source face unchanged and only performs makeup transfer on the facial texture. 
The key point is that we directly disentangle the makeup from the UV texture space rather than from the whole image with different shapes. 
It is proved to be an effective method for handling large pose and expression variations.

In the following, we introduce our makeup transfer flow in the texture space. 
We utilize an encoder-bottleneck-decoder architecture as the backbone. 
The encoder applies two convolutional layers to distill texture features $F_{src}$ from the source texture $T_{src}$. 
Then, in the bottleneck part, we transfer the makeup to the source features $F_{src}$ by concatenation. 
Finally, sending the combined features to the decoder to produce the transferred texture $T_{t}$. 
To handle the shadow and occlusion problem and obtain cleaner makeup representation, a reference encoder $E_{ref}$ and two attentive modules are introduced. 
In particular, the reference encoder $E_{ref}$ has the same structure as the encoder $E_{src}$, but they do not share parameters. 
Two attentive modules are designed to extract clean makeup features from the reference texture $T_{ref}$ and perform makeup transfer accurately. 
\vspace{1ex}

{\bf Flip Attention Module (FAM).}
When there is shadow and occlusion appearing on the reference image, e.g.\ , the hand covering the face, 
the extracted texture $F_{ref}$ will be contaminated and thus not accurate.
Flip Attention Module (FAM) is designed to adaptively fix imperfect regions in feature maps of the reference texture $T_{ref}$. 
Without using any manual annotations, FAM learns to locate the shadowed and occluded regions in a fully unsupervised way. 

Given the imperfect reference feature, FAM first generates an attentional mask $M$ through convolutions and a sigmoid operation, which has the same width and height with $F_{ref}$. 
The attentional mask $M$ learns the makeup quality of different regions and gives higher confidence to good quality regions and lower confidence to lower quality regions. 
For example, the facial regions which are occluded by the hand may have lower attentional weights than other regions.
 
Furthermore, inspired by \cite{Wu_2020_CVPR}, we consider the symmetry of the UV texture of faces and apply flipping operation to help repair the low-confidence areas. 
By doing this, we are assuming that facial makeup has a horizontal-symmetry structure, while contaminations are at random and not symmetrical, which is very frequently the true case in the real world.
Therefore, the repaired reference feature $\hat{F}_{ref}$ is obtained by, 

\begin{equation}
  \hat{F}_{ref}=M \otimes F_{ref} \oplus \left(1-M\right) \otimes F_{ref}^{\prime} \text{,}
  \label{con:FAM}
\end{equation}
where $\otimes$ denotes element-wise multiplication, $\oplus$ denotes element-wise addition.
The values of $M$ are between 0 and 1. 
\vspace{1ex}

{\bf Makeup Transfer Module (MTM).}
The refined reference feature $\hat{F}_{ref}$ is taken as the input of the Makeup Transfer Module (MTM). 
MTM is a spatial attention module \cite{wang2018non,zhang2019self,deng2020reference}, which combines the source feature and the reference feature for precisely transferring the corresponding face makeup of the reference image to the source image. 
The proposed structure is shown in the lower right part of Figure~\ref{fig:pipeline}.
Since both features are well-aligned in the UV space, each position across the network has the same semantic meaning. 
Therefore, we directly propose an attention map from the source feature to ask for the makeup information of the reference feature. 
We calculate the attention map from the source feature by $F_{a}= softmax(F_{p}F_{q})$, where $F_{p}$ and $F_{q}$ are features obtained by $1 \times 1$ convolution. 
To extract the makeup feature of the corresponding position, the reference feature is embedded by multiplying the attention map, denoted as $F_{m}=F_{a} \otimes \hat{F}_{ref}$. 
Finally, the source feature and the reference texture are concatenated to the decoder for outputting the transfer results. 
Such a design can effectively transfer the makeup from the reference feature into the source feature, which produces a more accurate makeup transfer result. 
\subsection{Loss Functions}
\label{subsection:loss}


{\bf Makeup loss.}
Similar to \cite{jiang2020psgan}, we also apply the makeup loss proposed by \cite{li2018beautygan} to add supervision for the makeup similarity. 
The makeup loss consists of three local color histogram loss on lips, eye shadow, and face, respectively.
We employ histogram matching between $I_{t}$ and $I_{ref}$, denotes as $HM(I_{t} \otimes M_{item},I_{ref} \otimes M_{item})$. 
In specific, $M_{item}$ is the local region obtained by a face parsing model, $item \in \{lips, eye, face\}$. 
As a kind of ground truth, $HM(I_{t} \otimes M_{item},I_{ref} \otimes M_{item})$ restricts the output image and the reference image to have similar makeup style in the locations of $M_{item}$.

The local histogram loss is formulated as 
\begin{equation}
	\begin{aligned}
\mathcal{L}_{item}=\left\|HM(I_{t} \otimes M_{item},I_{ref}\otimes M_{item}) -I_{t}\otimes M_{item}\right\|_{2} \text{.}
	\end{aligned}
\end{equation}

The overall makeup loss is formulated as
\begin{equation}
	\begin{aligned}
\mathcal{L}_{makeup}=\lambda_{1}\mathcal{L}_{lips}+\lambda_{2}\mathcal{L}_{eye}+\lambda_{3}\mathcal{L}_{face} \text{,}
	\end{aligned}
	\label{con:makeuploss}
\end{equation}

{\bf Perceptual loss.}
Perceptual loss is defined by the distance of the high-dimensional features between two images. 
In order to capture facial texture details and improve visual quality, we adopt it in the texture space to measure the distance between the generated texture $T_{t}$ and the source texture $T_{src}$. 
Here, we utilize a pre-trained $VGG$-16 model and take $relu$-4-1 outputs as the VGG network features. 

The perceptual loss can be expressed as follows: 

\begin{equation}
	\begin{aligned}
\mathcal{L}_{per}=\left\|VGG(T_{t})-VGG(T_{src})\right\|_{2} \text{.}
	\end{aligned}
\end{equation}

{\bf Cycle consistency loss.}
Since there are generally no paired ground-truth images for supervising the makeup transfer. 
We introduce the cycle consistency loss proposed by \cite{zhu2017unpaired} in the texture space, 
which is adapted to convert images back and forth between the makeup domain and the non-makeup domain. 
Here, it is defined as pixel-level L-1 distances between the reconstructed texture and the input texture. 
It is formulated as 

\begin{equation}
  \begin{aligned}
  \mathcal{L}_{cycle} &=\|G(G(T_{src}, T_{ref}), T_{src})-T_{src}\|_{1} \\
  &+\|G(G(T_{ref}, T_{src}), T_{ref})-T_{ref}\|_{1} \text{.}
  \end{aligned}
\end{equation}

{\bf Adversarial loss.}
To further improve the fidelity of the generated results, we use adversarial losses during the training. 
We introduce two domain discriminators $D_{tex}^{S}$, $D_{tex}^{R}$ and one real-fake discriminator $D_{img}$. 
$D_{tex}^{S}$, $D_{tex}^{R}$ are introduced for the generated UV texture maps and $D_{img}$ is for the rendered face. 
Specifically, $D_{tex}^{S}$ aims to distinguish the generated texture from the source texture domain $S$, $D_{tex}^{R}$ aims to distinguish the generated texture from the reference texture domain $R$: 

They are formulated as: 
\begin{equation}
	\begin{aligned}
  \mathcal{L}_{adv}^{D_{tex}}=&-\mathbb{E}\left[\log D_{tex}^{S}(T_{src})\right]-\mathbb{E}\left[\log D_{tex}^{R}(T_{ref})\right] \\
  &-\mathbb{E}\left[\log \left(1-D_{tex}^{S}(G(T_{ref}, T_{src}))\right)\right] \\
  &-\mathbb{E}\left[\log \left(1-D_{tex}^{R}(G(T_{src}, T_{ref}))\right)\right] \text{,}\\
  \mathcal{L}_{adv}^{G_{tex}}=&-\mathbb{E}\left[\log \left(D_{tex}^{S}(G(T_{ref}, T_{src}))\right)\right] \\
  &-\mathbb{E}\left[\log \left(D_{tex}^{R}(G(T_{src}, T_{ref}))\right)\right] \text{.}
  \end{aligned}
\end{equation}

$D_{img}$ is trained to tell whether the generated outputs are real or fake, formulated as: 
\begin{equation}
	\begin{aligned}
  \mathcal{L}_{adv}^{D_{img}}=&-\mathbb{E}\left[\log D_{img}(I_{src})\right]-\mathbb{E}\left[\log D_{img}(I_{ref})\right] \\
  &-\mathbb{E}\left[\log \left(1-D_{img}(R(S_{ref},G(I_{ref}, I_{src})))\right)\right] \\
  &-\mathbb{E}\left[\log \left(1-D_{img}(R(S_{src},G(I_{src}, I_{ref})))\right)\right] \text{,}\\
  \mathcal{L}_{adv}^{G_{img}}=&-\mathbb{E}\left[\log \left(D_{img}(R(S_{ref},G(I_{ref}, I_{src})))\right)\right] \\
  &-\mathbb{E}\left[\log \left(D_{img}(R(S_{src},G(I_{src}, I_{ref})))\right)\right]\text{.}
  \end{aligned}
\end{equation}


{\bf Total loss.}
The full objective function can be expressed as follows:
\begin{equation}
  \begin{aligned}
  \mathcal{L}_{G} &= \lambda_{a}\mathcal{L}_{adv}^{G_{tex}} + \lambda_{a}\mathcal{L}_{adv}^{G_{img}} + \lambda_{m}\mathcal{L}_{makeup} + \lambda_{c}\mathcal{L}_{cycle} + \lambda_{p}\mathcal{L}_{per}  \text{,}\\
  \mathcal{L}_{D} &= \lambda_{a}\mathcal{L}_{adv}^{D_{tex}} + \lambda_{a}\mathcal{L}_{adv}^{D_{img}} \text{.}
  \end{aligned}
  \label{con:totalloss}
\end{equation}

\section{EXPERIMENTS}
\vspace{1ex}

\subsection{Implementation Details}

Makeup Transfer (MT) dataset is currently the largest facial makeup dataset proposed by \cite{li2018beautygan}. 
It gathers a total of 3,834 female images, with 1,115 non-makeup images and 2,719 makeup images. 
It contains a series of makeup styles from subtle to heavy, such as Korean makeup style, flashy makeup style, and smoky-eyes makeup style. 
All the images are aligned using 68 facial landmarks and have the same spatial size $256 \times 256$. 
Following the train/test protocol of \cite{li2018beautygan}, we randomly choose 250 makeup images and 100 non-makeup images as our test dataset for performance evaluation. 
The rest images are used as our training set. 
More recently, PSGAN \cite{jiang2020psgan} collected a Makeup-Wild dataset that consists of facial images with different poses and expressions. 
It contains a total of 772 female images, with 369 non-makeup images and 403 makeup images. 
We also test on the Makeup-Wild dataset to verify the robustness of our method under large pose and expression.

Our SOGAN is trained on 1 NVIDIA RTX 2080Ti GPU, with PyTorch3D (v0.2.0) \cite{paysan20093d} and its dependencies.
For the optimizer, we choose ADAM~\cite{kingma2014adam} with $\beta_{1}$ = 0, $\beta_{2}$ = 0.9.
We use a batch size of 1 and a constant learning rate of 0.0002.
The trade-off parameters in Eq.~\ref{con:makeuploss} are set as $\lambda_{1}=\lambda_{2}=1, \lambda_{3}=0.1$.
The trade-off parameters in Eq.~\ref{con:totalloss} are set as $\lambda_{a}=\lambda_{m}=1, \lambda_{c}=10, \lambda_{p}=5e-3$.
\vspace{1ex}

\subsection{Comparison with Other Methods}
In the following, we show quantitative and qualitative results of our method to verify its effectiveness. 

\subsubsection{Results under Shadow and Occlusion.}

\begin{figure}[H]
	\begin{center}
		\includegraphics[width=1.0\linewidth]{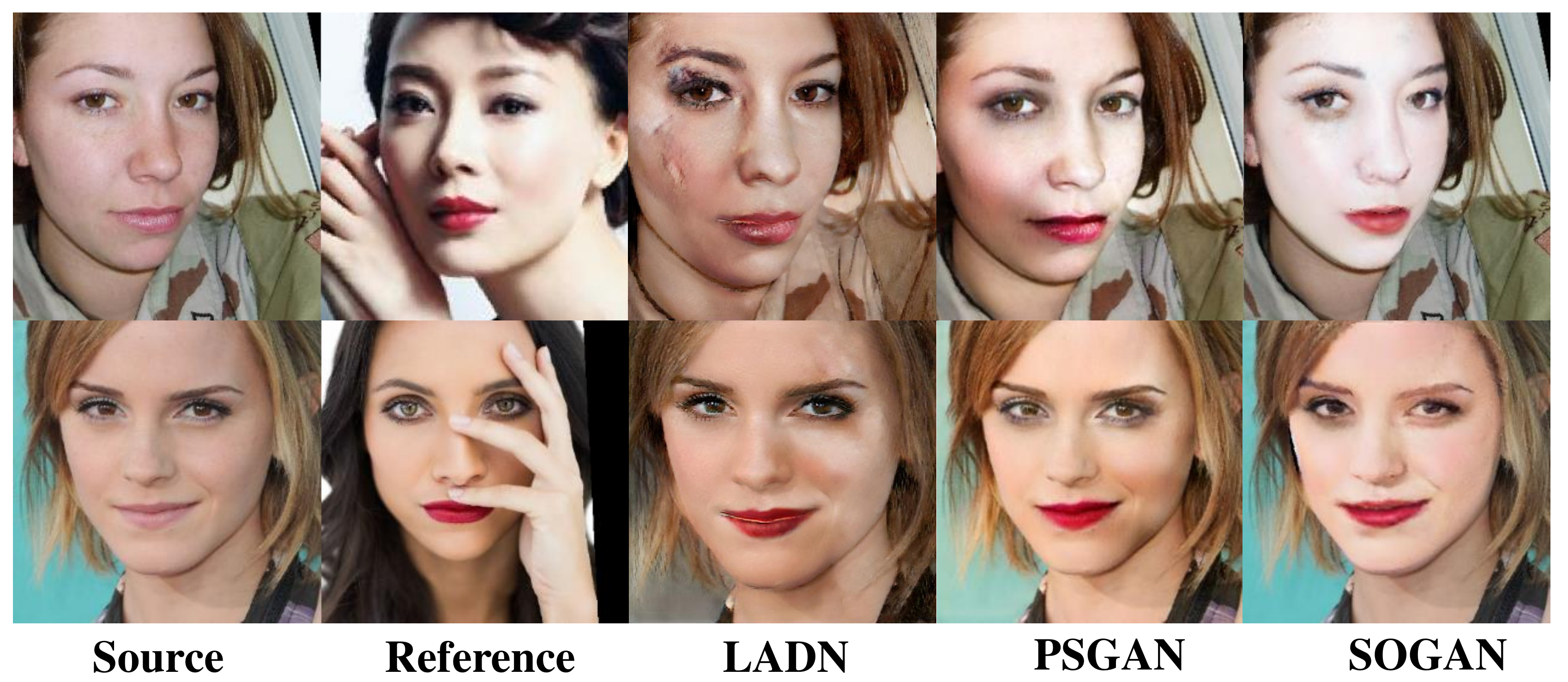}
	\end{center}
	\caption{Visual comparisons with state-of-the-art methods when dealing with images under shadow and occlusion.}
	\label{fig:occlusion}
\end{figure}

We conduct a comparison on shadowed and occluded images with previous state-of-the-art methods, LADN \cite{gu2019ladn} and PSGAN \cite{jiang2020psgan}.  
For fairness, we use their released code and pre-trained model. 
The comparison results are shown in Figure~\ref{fig:occlusion}. 
For LADN results, we can see obvious occlusion marks and ghosting artifacts in the transferred images. 
The quality of PSGAN is better than that of LADN. 
However, it still has trouble repairing the occluded areas, since it uses pixel-to-pixel makeup transfer in the 2D image space, which is not as effective as our 3D-aware method. 
Our SOGAN repairs the shadowed and occluded regions in the transferred images by applying the proposed FAM module. 

\subsubsection{Results under Large Pose and Expression}
\begin{figure}[H]
	\begin{center}
		\includegraphics[width=1.0\linewidth]{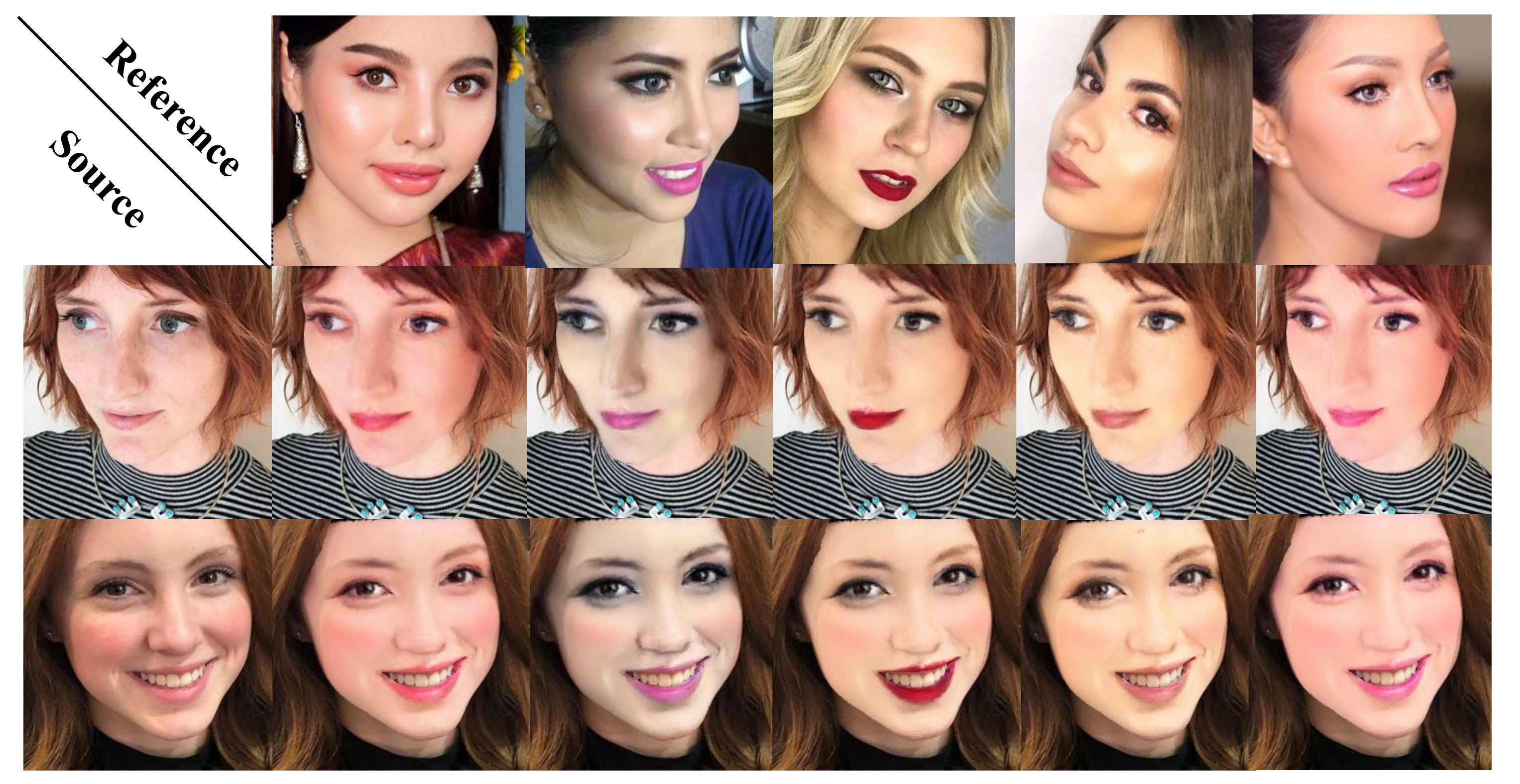}
	\end{center}
	\caption{Illustrations of our makeup transfer results under large pose and expression variations. The first column is the source images and the first row is the reference images. Our method performs well on all transferred images.}
	\label{fig:poseandexpression}
\end{figure}

To examine the robustness under large pose and expression, we test our method on the Makeup-Wild dataset. Multiple samples with different poses and expressions are selected. 
These sample results are shown in the Figure~\ref{fig:poseandexpression}. The first column is the input source images, the first row is the input reference images, any pair of source and reference images have different poses and expressions, different makeup styles, and different face sizes.
Our method performs well on all transferred images even if source and reference images show different variations. 

\begin{figure}[htb]
	\begin{center}
		\includegraphics[width=1.0\linewidth]{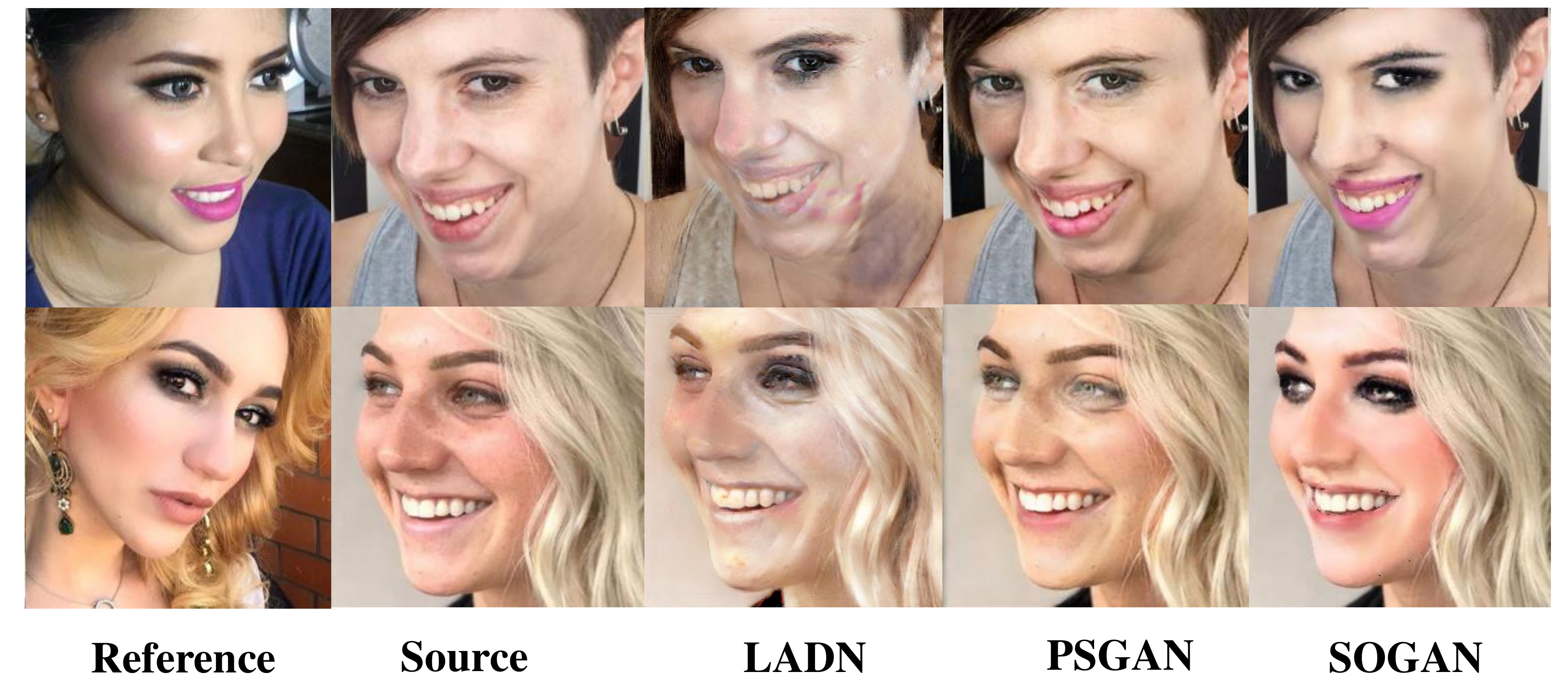}
	\end{center}
	\caption{Visual comparisons with state-of-the-art methods when dealing with images with different poses and expressions on Makeup-Wild dataset.}
	\label{fig:largepose comparsion}
\end{figure}

We then conduct a comparison on the Makeup-Wild dataset with previous state-of-the-art methods, LADN\cite{gu2019ladn} and PSGAN\cite{jiang2020psgan}. 
For a fair comparison, we use their provided source code and pre-trained model. Note that we only train our network using the training part of the MT dataset. 
The comparison results are shown in Figure~\ref{fig:largepose comparsion}.
LADN transfers the makeup from the reference image to the wrong region of the source image since it overfits on frontal faces with netural expression.
For example, in the first row, LADN transfers the lipstick on the reference face to the skin region of the source face.
PSGAN is robust to variations in pose and expression due to its spatial attention module.
It accurately transfers the makeup from the reference image to the corresponding position in the source image. 
However, in the first row, when the reference image has shadow caused by a large pose, PSGAN shows an imperfect makeup transfer result. 
Since our SOGAN transfers the makeup in the UV space, the pose and expression variations are explicitly normalized. 
Our method can naturally achieve impressive transferred results under large pose and expression.
It is shown that we achieve accurate makeup transfer results on both images.

\vspace{1ex}
\subsubsection{Results on Frontal Faces in Neutral Expressions.}
\begin{figure*}[htb]
	\begin{center}
		\includegraphics[width=1\linewidth]{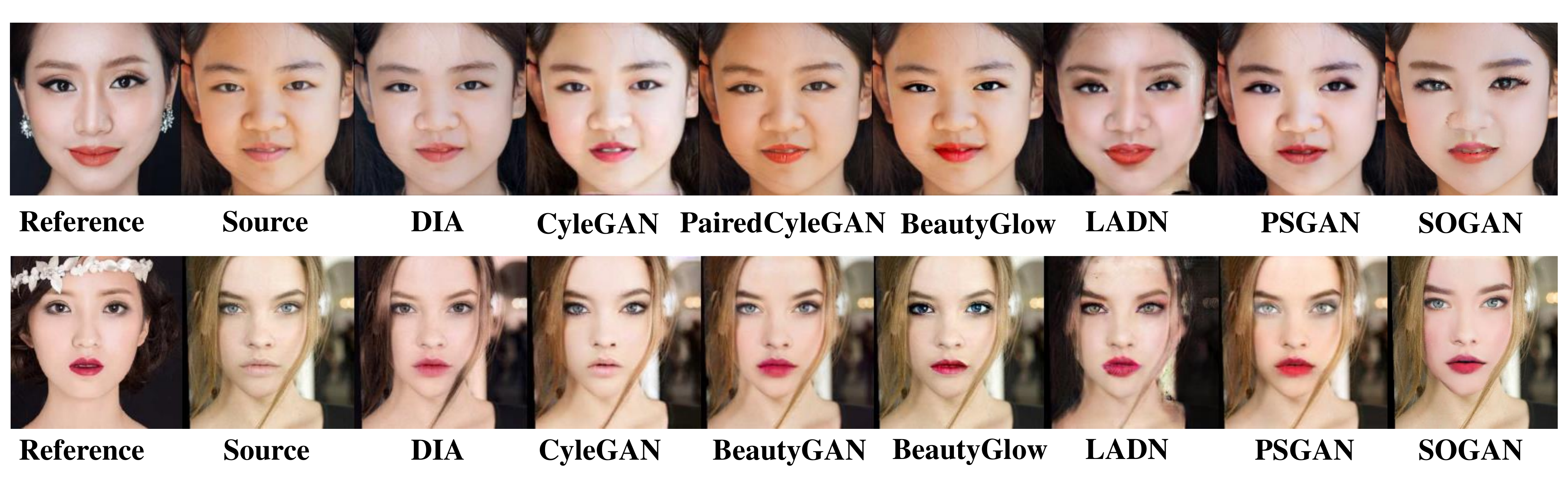}
	\end{center}
	\vspace{-1ex}
	\caption{Qualitative Comparisions with state-of-the-art methods on frontal faces with neutral expressions. Our method can generate high-quality makeup transfer results, especially with more accurate eye makeup.}
	\label{fig:qualitative comparison}
\end{figure*}

We conduct a comparison of SOGAN with other previous state-of-the-art methods on frontal faces in neutral expressions. 
In particular, BeautyGAN\cite{li2018beautygan}, BeautyGlow\cite{chen2019beautyglow} and PairedCycleGAN\cite{chang2018pairedcyclegan} do not release their code and pre-trained model.
For a fair comparison, we crop the results from the corresponding papers. 
The results are shown in the Figure~\ref{fig:qualitative comparison}.
For DIA results, it transfers the style of the entire image. 
Thus, the background of the reference image is also transferred to the source image, producing inappropriate makeup transfer results.
Since CycleGAN transfers a general makeup style rather than makeup details, the results of CycleGAN lose makeup details.
The quality of BeautyGAN is better than that of DIA and CycleGAN. 
However, since BeautyGAN transfers in the whole image, the color of the hair and the background of the image are changed.
For the results of BeautyGlow, they do not have the same color of lipstick as the reference, but they preserve the background better. 
LADN produces blurry and unrealistic transfer results and changes the source background. 
PSGAN achieve nice transfer results. However, it does not accurately transfer the eye shadow from the reference image to the source image.
Compared with previous methods, SOGAN transfers the long eyelashes in the reference image to the source image, which is considered as a part of eye makeup. 

\vspace{1ex}
In summary, SOGAN performs well on frontal images in neutral expressions, further on different pose and expression, shadow and occlusion situations.

\subsubsection{Quantitative Comparison.}
We conduct user studies to compare the performance of the proposed method with CycleGAN\cite{zhu2017unpaired}, LADN\cite{gu2019ladn} and PSGAN\cite{jiang2020psgan}. 
Firstly, we select 20 source images and 20 reference images that have clean makeup from the MT test set and obtain generated images for each method. 
Randomly giving 20 generated images to 20 participants, we let them select the best from four methods considering image realism and the similarity with reference makeup styles.
The first row of Table~\ref{tab:table1} shows the averaged selection percentage for each method when inputting clean reference makeup. 
Our method and PSGAN outperform other methods with a large margin. 

\vspace{1ex}
We then conduct another user study to test the effectiveness of our method in handling contaminated makeup images. 
We select 20 source images and 20 contaminated reference images from the MT test set and obtain images for each method. 
Similarly, randomly selecting 20 generated images to 20 participants, we let them select the best from four methods considering image realism and the degree of shadow and occlusion removal.
As can be observed in Table~\ref{tab:table1}, 
when going from clean reference images as inputs to contaminated reference images as inputs, the performance of our SOGAN improves by 5\%, while the performance of PSGAN reduces by 10\%, and other methods still have lower quality. 
our method achieves the best performance when handling contaminated reference makeup images. 

Note that all the generated images are shown in random order for a fair comparison. Besides, we do not collect the privacy information of the evaluators and all questionnaires are anonymous.
\begin{table}[H]
	\centering
	\large
	\begin{tabular}{cccccc}
		\hline  Test Set & SOGAN & PSGAN & LADN & CycleGAN \\
		\hline  Clean Set &$\mathbf{51.50\%}$ & $41.33\%$ & $5.17\%$ & $2.00\%$ \\
		\hline  SO Set & $\mathbf{56.25\% }$ & $31.00\% $ & $9.75\%$ & $3.00\%$\\
		\hline
		\end{tabular}
		\vspace{2ex}
		\caption{Average ratio selected as best (\%). The "Clean Set" row represents testing images that all have clean makeup, while the " SO Set" row represents testing images that are all under shadow or occlusion conditions.}
\label{tab:table1}
\end{table}

\subsection{Partial and Interpolated Makeup Transfer}

\begin{figure}[htb]
	\begin{center}
		\includegraphics[width=1.0\linewidth]{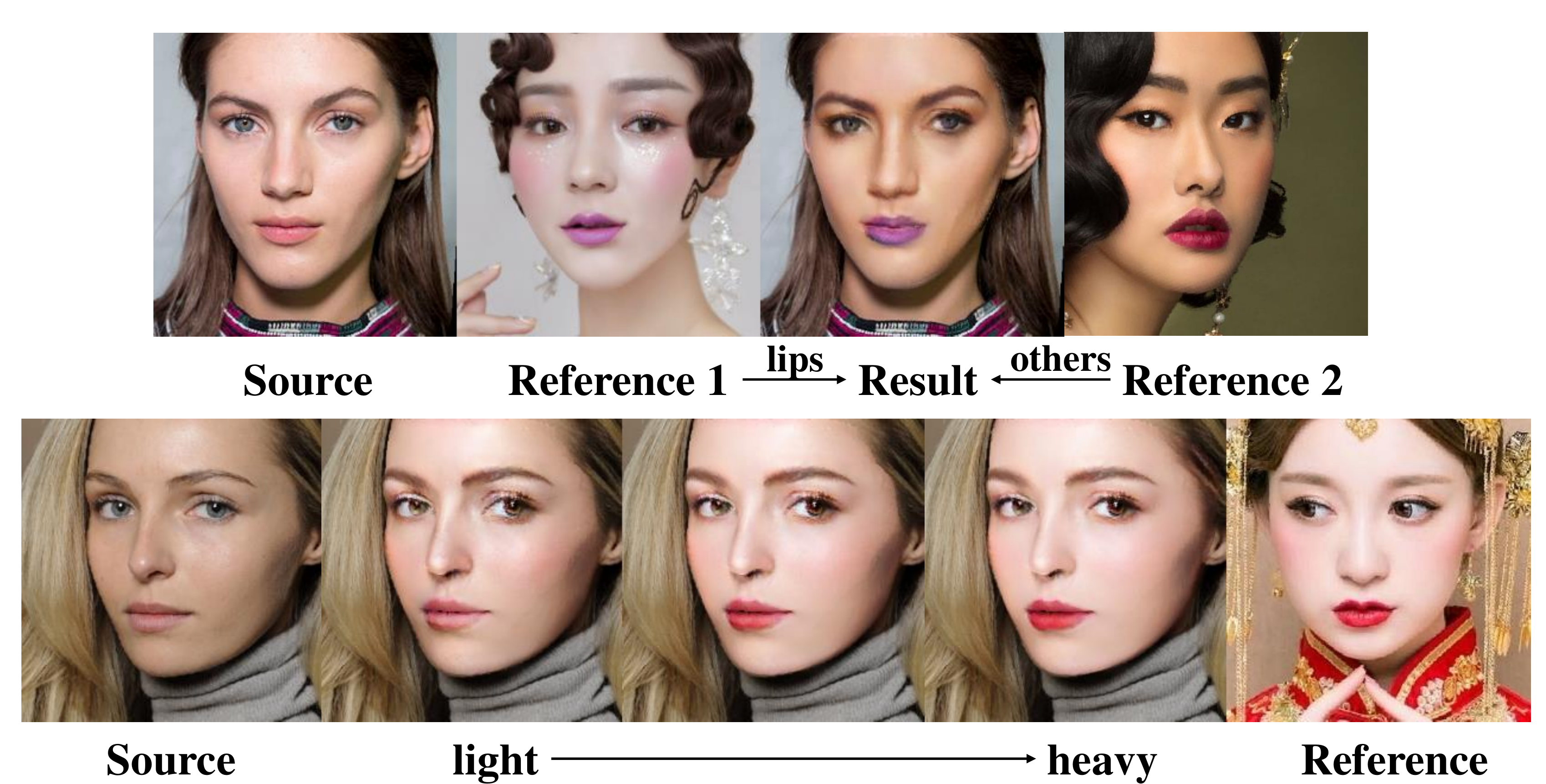}
	\end{center}
	\caption{Illustrations of partial makeup transfer and shade controllable makeup transfer. The first row shows the result of transferring the lipstick from reference 1 and other makeup from reference 2. The second row shows the makeup transfer effect from light to heavy.}
	\label{fig:partialandlighttoheavy}
\end{figure}

We can perform partial transfer during testing since we design the Makeup Transfer Module (MTM) in a spatial-aware way.
In order to achieve partial makeup transfer, the proposed MTM are trained using a face parsing map, inspired by \cite{jiang2020psgan}. 
After training, we can achieve partial makeup transfer by the guidance of the source face parsing mask. 
As shown in the first row of Figure~\ref{fig:partialandlighttoheavy}, we use the source face parsing mask of lip $M_{lips}$ to extract the lipstick makeup from reference 1 and other parts of the mask to extract other makeup from reference 2.  
As a result, we produce a new mixed makeup style.

we can also control the shade of makeup transfer by introducing a weight $w$ of the attention map $F_{a}$ and let $F_{m}= w \times F_{a} \otimes \hat{F}_{ref}$.
As shown in the second row of Figure~\ref{fig:partialandlighttoheavy}, we set the weight $w$ from 0 to 1 and achieve the makeup transfer from light to heavy. 

\begin{figure*}[htb]
	\begin{center}
		\includegraphics[width=1.0\linewidth]{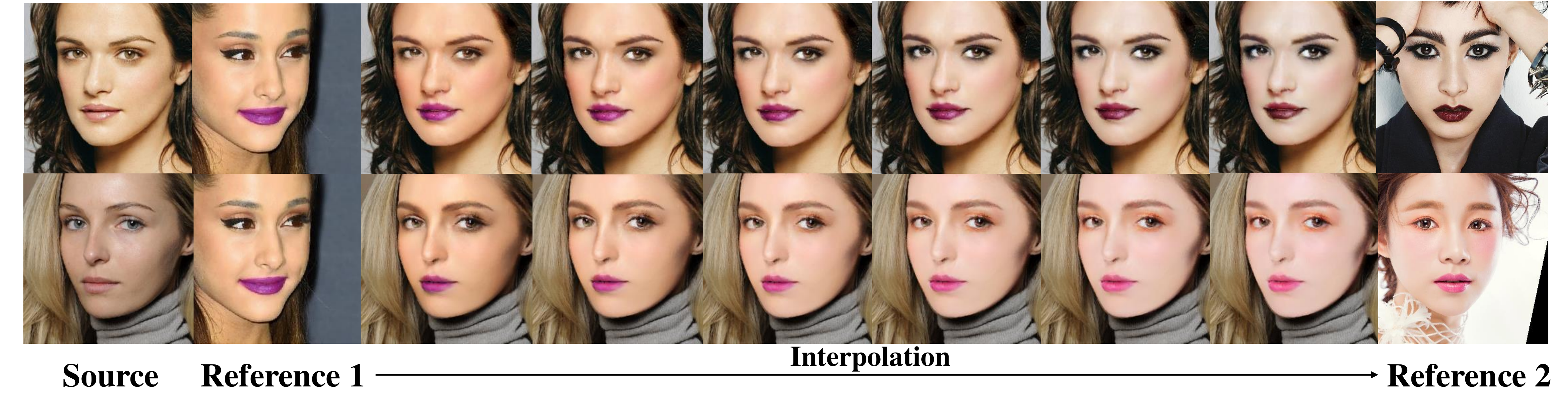}
	\end{center}
	\vspace{-2ex}
	\caption{Results of facial makeup interpolation. The makeup styles of the interpolated images continuously transfer from reference 1 to reference 2 while simultaneously preserving the source information.}
	\label{fig:interpolation}
\end{figure*}

In order to achieve facial makeup interpolation, we compute the interpolated feature $F_{m}^{i} = w \times F_{a} \otimes \hat{F}_{ref1} + (1 - w) \times F_{a} \otimes \hat{F}_{ref2}$, where $\hat{F}_{ref1}$ and $\hat{F}_{ref2}$ are features extracted from the referece 1 and the reference 2. 
As shown in Figure~\ref{fig:interpolation}, we achieve natural and realistic interpolated results from reference 1 to reference 2, by setting the weight $w$ from 0 to 1 in intervals of 0.2.

\subsection{Analysis of Key Modules}

Our network mainly consists of the FAM module and the MTM module, which work together to obtain nice and robust transfer results. 
We conduct some analysis and ablation studies to examine the effectiveness of these modules. 

\subsubsection{Effectiveness of Flip Attention Module (FAM)}
\begin{figure}[htb]
	\begin{center}
		\includegraphics[width=1.0\linewidth]{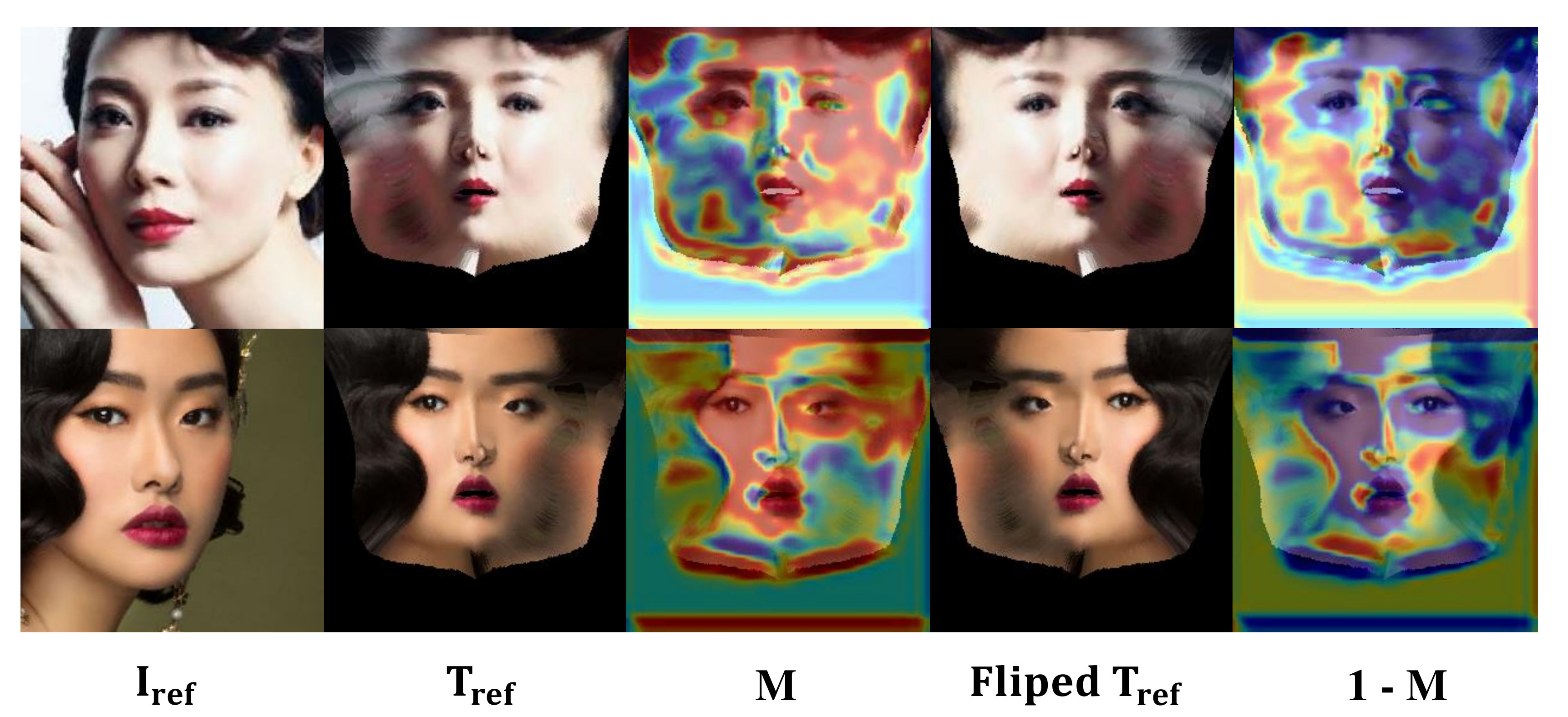}
	\end{center}
	\vspace{-2ex}
	\caption{Visualizations of the attentional mask $M$ in the FAM module. $M$ learns the makeup quality of different regions and gives higher confidence to cleaner regions and lower confidence to contaminated regions. }
	\label{fig:visulization}
\end{figure}

In SOGAN, the FAM module is proposed to repair shadowed and occluded regions in the reference makeup images and obtain cleaner makeup representation. 
It achieves this goal by applying flipping operation and attention fusion. 
We verify the effectiveness of the FAM module in this part. 
As shown in Figure~\ref{fig:visulization}, the first two columns are the input reference images and the corresponding UV texture maps, 
they have shadow and occlusion in some regions, which will cause ghosting artifacts in the generated images. 
By applying FAM, the contaminated regions are located by the attentional mask $M$. 
For example, in the first row, the attentional mask $M$ gives high confidence to the clean regions and low confidence to the shaded area on the bottom left of the face. 
In the second row, for the occlusion of hair and the self-occlusion of the nose caused by the face pose, the attentional mask $M$ also gives accurate confidence. 
Guided by $M$, the FAM module can use information from both original and flipped texture map to repair the contaminated regions by Eq.~\ref{con:FAM}. 

In Figure~\ref{fig:ablation}, we further show the effectiveness of the FAM module through ablation study.
The first column and the second column are the input reference images and source images, respectively. 
In the first row, the reference image has occlusion on the left face boundary. 
As a result, the transferred image shows occlusion artifacts at the same regions without FAM. 
Similarly, in the second row, the generated face without FAM shows obvious shadow marks on the right regions, which is caused by the shadowed makeup in the reference image. 
By introducing FAM, it can adaptively fix the imperfect reference makeup in an unsupervised manner. 

\subsubsection{Effectiveness of Makeup Transfer Module (MTM)}
\begin{figure}[htb]
	\begin{center}
		\includegraphics[width=1.0\linewidth]{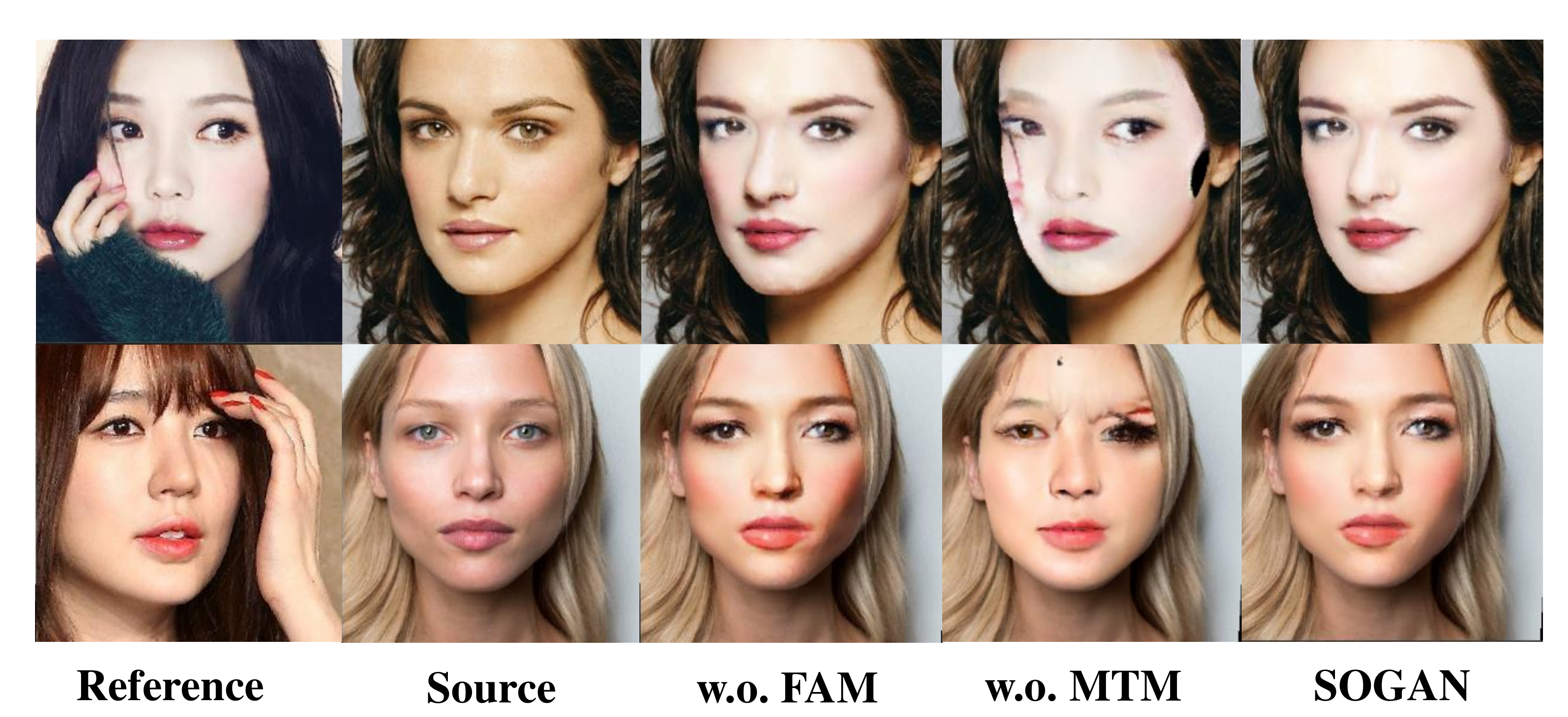}
	\end{center}
	\vspace{-2ex}
	\caption{Comparisions under different ablation configurations, including "w / o FAM module" and "w / o MTM module".}
	\label{fig:ablation}
\end{figure}

The effectiveness of the proposed MTM module is also shown in the Figure~\ref{fig:ablation}.
The generated images change the original identity and show severe ghosting artifacts without MTM. 
By applying the MTM module, accurate makeup transfer can be achieved. 


\section{Conclusion}

In this paper, we propose a novel 3D-Aware Shadow and Occlusion Robust GAN (SOGAN). 
In order to push the makeup transfer task towards real-world usage, we try to solve the problem of undesired shadow and occlusion in the reference images.
We disentangle the shape and texture of input images by fitting a 3D face model and then map the texture to the UV space.
While keeping the shape of the source image unchanged, the makeup is applied from the reference texture to the source texture in the UV space.
To repair the shadowed and occluded regions in the reference images, we design a Flip Attention Module (FAM). 
To achieve more accurate makeup transfer, we design a Makeup Transfer Module (MTM). 
These modules work together to solve the problem and achieve promising makeup transfer results. 

\begin{acks}
	The authors would like to thank Chang Yu, Xiangyu Zhu from NLPR, CASIA for their helpful discussions and suggestions on this work.
	This work was supported by National Natural Science Foundation of China under Grants No. 61772529, 61972395, 61902400, U19B2038 and Beijing Natural Science Foundation under Grant No. 4192058.
\end{acks}
\bibliographystyle{ACM-Reference-Format}
\balance
\bibliography{sample-base}

\appendix

\end{document}